\documentclass[10pt,twocolumn,letterpaper]{article}

\usepackage[final]{cvpr}      
\usepackage{times}
\usepackage{epsfig}
\usepackage{graphicx}
\usepackage{amsmath}
\usepackage{amssymb}
\usepackage{enumitem}
\usepackage{tikz}
\usetikzlibrary{arrows.meta,positioning}
\usepackage[breaklinks=true,colorlinks,allcolors=blue]{hyperref}

\def\confName{CVPR}
\def\confYear{2026}

\definecolor{cblue}{HTML}{9BB8D3}
\definecolor{cred}{HTML}{C0392B}
\definecolor{cgreen}{HTML}{27AE60}
\definecolor{cdark}{HTML}{2C3E50}
\newcommand{\best}[1]{\textbf{#1}}

\begin{document}

\title{Perception First: A Frontier Native-Video Model with Self-Consistency\\
for Implicit Video Question Answering}

\author{Ali Alavi\\
The Ohio State University\\
{\tt\small alavibajestan.1@osu.edu}
}
\maketitle

\begin{abstract}
We describe our submission to the VRR Challenge @ CVPR 2026, built on the \emph{ImplicitQA} /
\emph{VRR-QA} benchmark~\cite{implicitqa}: multiple-choice video question answering in which answers
are deliberately \emph{not} observable in any single frame and must be inferred from spatial layout,
motion, depth, viewpoint, causality, and social context across discontinuous frames of creative
video. We conduct a systematic, training-free study spanning open-source Video-LMMs
(Qwen2.5-VL~\cite{qwen25vl}, Qwen3-VL~\cite{qwen3vl}, InternVL3, Gemma-3, and the RL-tuned video
reasoners Video-R1~\cite{videor1} and VideoChat-R1.5~\cite{videochatr15}) and a battery of
inference-time strategies (chain-of-thought, question decomposition, describe-then-reason cascades,
audio transcripts, spatial state prompting, self-consistency~\cite{selfconsistency}, multi-model
ensembling, and category routing). Our central finding is that this benchmark is
\emph{perception-bound rather than reasoning-bound}: reasoning-side augmentations are
neutral-to-harmful, whereas base-model perceptual capability and lightweight test-time denoising are
the only reliable levers. A per-category error analysis localizes the difficulty to low-level
perception---relative depth, viewpoint, and counting are the hardest categories, while causal and
social reasoning are nearly solved---and a prompt that explicitly injects monocular depth cues to
attack the weakest category \emph{lowers} test accuracy by $5.8$ points, confirming that the model
needs a better \emph{percept}, not a better \emph{procedure}. The open-source ceiling we reach is
$58.5\%$ accuracy (Qwen3-VL-32B-AWQ with self-consistency). Our final system---a current
\emph{native-video} frontier model, Gemini~3.1~Pro~\cite{gemini}, with $5$-way self-consistency---attains
\best{81.18\%} average and \best{78.85\%} macro-average accuracy on the hidden test split, surpassing
the prior best ($80.85\%$) and approaching the non-expert human baseline ($83.0/85.6$). The pipeline is
inference-only and fully cached for exact reproduction.
\end{abstract}

\begin{figure*}[t]
\centering
\resizebox{0.94\textwidth}{!}{%
\begin{tikzpicture}[
  font=\small,
  box/.style={draw,rounded corners=3pt,align=center,minimum height=12mm,minimum width=22mm,line width=0.6pt},
  arr/.style={-{Latex[length=2.4mm]},line width=0.9pt}]
  \node[box,fill=cblue!25] (clip) {Trimmed\\video clip\\\scriptsize(native, +audio)};
  \node[box,fill=cred!15,right=11mm of clip] (model) {\textbf{Gemini 3.1 Pro}\\\scriptsize chain-of-thought};
  \node[box,fill=cblue!18,right=11mm of model,minimum width=28mm] (samp)
      {$k{=}5$ samples\\\scriptsize($T{=}0.7$)\\[1pt]\ttfamily A\ B\ A\ A\ A};
  \node[box,fill=cblue!18,right=10mm of samp] (vote) {Majority\\vote};
  \node[box,fill=cgreen!18,right=10mm of vote] (ans) {\textbf{Answer:}\\\textbf{A}};
  \draw[arr] (clip)--(model);
  \draw[arr] (model)--(samp);
  \draw[arr] (samp)--(vote);
  \draw[arr] (vote)--(ans);
  \node[below=2mm of model,font=\scriptsize\itshape,text=cdark] {native video $\gg$ frame sampling};
  \node[below=2mm of vote,font=\scriptsize\itshape,text=cdark] {self-consistency};
\end{tikzpicture}}
\caption{\textbf{Overview of the final system.} Each pre-trimmed clip is fed as \emph{native video}
to Gemini~3.1~Pro, which produces $k{=}5$ independent chain-of-thought answers at temperature $0.7$; a
majority vote yields the prediction. Native video ingestion supplies the perceptual capability the
benchmark demands, and self-consistency denoises the strong base---together reaching $81.2\%$ test
accuracy. Uploads and responses are cached, so re-runs are free.}
\label{fig:overview}
\end{figure*}
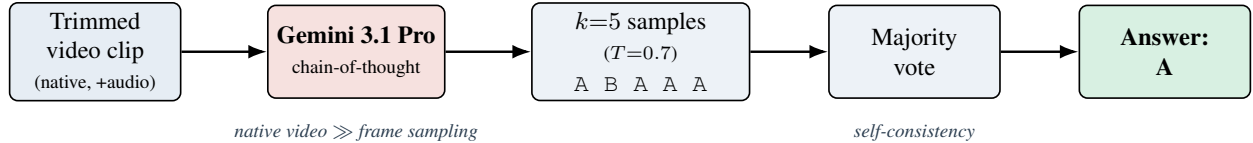

\section{Benchmark and Approach Overview}
\textbf{Benchmark.} ImplicitQA/VRR-QA~\cite{implicitqa} evaluates \emph{implicit} reasoning in
Video-LMMs: each multiple-choice question is constructed so that its answer cannot be read off any
single frame, but must be inferred across frames. The challenge exposes two phases: a
\textbf{Validation} split of $1{,}001$ QA pairs with \emph{public} labels (our development set) and a
hidden \textbf{Test} split of $172$ QA pairs that decides the final ranking. Questions span nine
reasoning categories; roughly $69\%$ are spatial (lateral, vertical, relative depth/proximity,
viewpoint). The number of options varies ($2$--$8$; the gold-letter prior favors B, so chance
$\approx 30\%$). Organizers provide pre-trimmed clips named by \texttt{question\_id} (mean duration
$16.6$\,s on test). Scoring uses \textbf{Average Accuracy} and category-wise \textbf{Macro-Average
Accuracy}.

\textbf{Approach.} We treat the task as \emph{inference $+$ test-time compute}: there is no
in-distribution training set, so gains must come from (i) base-model choice, (ii) how video is
presented to the model, (iii) prompting, and (iv) test-time aggregation (Fig.~\ref{fig:overview}). We
built a modular pipeline and used the labeled validation split to select every design decision before
committing it to the hidden test split. The hypothesis that emerged early---and that all ablations
confirmed---is that \emph{the bottleneck is perception, not reasoning search}.

\section{Architecture, Prompting, and System Design}
\subsection{Three-layer pipeline}
The codebase separates concerns behind stable interfaces (factory pattern $+$ abstract base classes):
\begin{itemize}[leftmargin=1.2em,itemsep=1pt]
  \item \textbf{Data layer:} loads QA pairs, resolves the trimmed clip per \texttt{question\_id},
  uniformly samples frames (cached), and optionally attaches a Whisper audio transcript.
  \item \textbf{Model layer:} a common \texttt{predict\_batch} interface implemented by (a) a
  vLLM~\cite{vllm}-backed multimodal model for open weights and (b) a Gemini API model.
  \item \textbf{Runner:} ties data and model via a fixed \texttt{QASample}/\texttt{Prediction}
  contract, scores against labels when present, and emits the submission JSON; a data-driven fallback
  to the majority class (``B'') guarantees a valid answer for any failed item.
\end{itemize}

\subsection{Serving}
Open models are served with vLLM; frames are passed as multi-image inputs. Because the cluster GPUs
are A100-40GB (no NVLink), tensor parallelism deadlocked on NCCL/shared-memory broadcast, so we run
\textbf{data-parallel} inference (one replica per GPU over a shard, then merge) and use 4-bit
\textbf{AWQ} quantization (the Marlin kernel works on Ampere, whereas FP8 does not) with eager
execution to fit large models on a single card. The final system uses \textbf{Gemini~3.1~Pro}
(\texttt{gemini-3.1-pro-preview}), which ingests the \emph{native video} (and audio) rather than
sampled frames; each clip is uploaded once and every response is cached to disk.

\subsection{Prompting and test-time compute}
The default prompt presents the question and (variable-count) options, requests chain-of-thought, and
constrains the final line to \texttt{Answer: <letter>}; a robust parser handles
\texttt{<think>}/\texttt{<answer>} formats. We also ablate a \emph{reasoner} template, a \emph{spatial}
``describe-the-layout-first'' template, a two-stage \emph{cascade} (blind description $\rightarrow$
answer), and a three-stage \emph{decomposition} (atomic sub-questions $\rightarrow$ answer over frames
$\rightarrow$ aggregate). \textbf{Self-consistency}~\cite{selfconsistency} samples $k$ reasoning paths
at temperature $T$ and takes the majority vote; this is the one component that transferred from
validation to test on a strong base.

\section{Training Details and Implementation Setup}
\textbf{Training.} \emph{None.} The system is entirely training-free; gains come from base-model
selection, video presentation, prompting, and test-time aggregation. \textbf{Setup.} Open-model
experiments ran on NVIDIA A100-40GB GPUs via SLURM with vLLM~0.18.1 (PyTorch~2.10, Transformers~5.7)
under data-parallel sharding; audio uses \texttt{faster-whisper}. The frontier system uses
\texttt{google-genai}~2.7 against \texttt{gemini-3.1-pro-preview} and needs only CPU $+$ network.
Final configuration: native video input; self-consistency $k{=}5$, $T{=}0.7$,
\texttt{max\_output\_tokens}$=2048$. Open baselines used $16$ uniform frames at $448$px. The full
Gemini run cost ${\approx}\$11.8$; responses and uploads are cached, so re-runs make zero API calls.

\section{Results}
\begin{figure}[t]
\centering
\includegraphics[width=\linewidth]{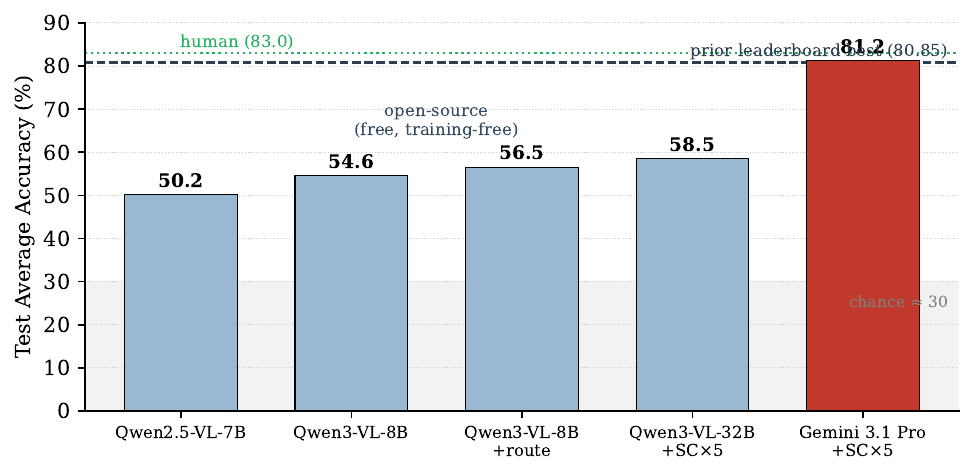}
\caption{\textbf{Test average accuracy across systems.} The training-free open stack climbs from
$50.2$ to a $58.5$ ceiling; the frontier native-video model with self-consistency reaches $81.2$,
above the prior best (dashed) and approaching non-expert human performance (dotted).}
\label{fig:results}
\end{figure}

Table~\ref{tab:main} reports validation ($n{=}1001$) and test ($n{=}172$) accuracy for key systems;
Fig.~\ref{fig:results} visualizes the test progression. Test is consistently $\sim$$5$--$8$ points
\emph{easier} than validation for a fixed system, so we use validation for all selection.

\begin{table}[t]
\centering\small
\setlength{\tabcolsep}{4pt}
\begin{tabular}{@{}l cc cc@{}}
\toprule
& \multicolumn{2}{c}{\textbf{Val}} & \multicolumn{2}{c}{\textbf{Test}} \\
\cmidrule(lr){2-3}\cmidrule(lr){4-5}
\textbf{System} & Avg & Mac & Avg & Mac \\
\midrule
Qwen2.5-VL-7B (CoT)                          & 45.8 & 48.7 & 50.2 & 47.3 \\
Qwen3-VL-8B (CoT)                            & 46.7 & 49.9 & 54.6 & 52.3 \\
\;\;$+$ decomposition                        & 46.7 & 50.3 & 56.2 & 53.1 \\
\;\;$+$ route (single$\leftrightarrow$dec.)  & 47.5 & 51.3 & 56.5 & 54.8 \\
Qwen3-VL-32B-AWQ (CoT)                        & 48.7 & 52.5 & 55.9 & 56.2 \\
\;\;$+$ self-consistency $(k{=}5)$            & \best{51.3} & \best{54.5} & 58.5 & 56.2 \\
\midrule
\best{Gemini 3.1 Pro $+$ SC $(k{=}5)$}        & --- & --- & \best{81.2} & \best{78.9} \\
\midrule
\textit{Prior best}                          &      &      & \textit{80.9} & --- \\
\textit{GPT-o3~\cite{implicitqa}}            & 64.1 & 68.6 &      & \\
\textit{Human (non-expert)~\cite{implicitqa}}& 83.0 & 85.6 &      & \\
\bottomrule
\end{tabular}
\caption{Average / Macro accuracy (\%). Open-source peaks at $58.5$ on test; the frontier native-video
model with self-consistency reaches $81.2$, surpassing the prior best of $80.9$.}
\label{tab:main}
\end{table}

\section{Ablation Studies and Observations}
All ablations are on the labeled validation split (full $1{,}001$ questions unless a $400$-question
subset is noted). Table~\ref{tab:abl} summarizes the decisive ones.

\begin{table*}[t]
\centering\small
\setlength{\tabcolsep}{8pt}
\renewcommand{\arraystretch}{1.1}
\resizebox{\textwidth}{!}{%
\begin{tabular}{@{}l l c c l@{}}
\toprule
\textbf{Lever} & \textbf{Setting} & \textbf{Val Avg} & \textbf{Effect} & \textbf{Takeaway}\\
\midrule
Base scale (CoT, greedy) & Qwen2.5-VL-7B $\to$ Qwen3-VL-32B-AWQ & $45.8\to48.7$ & $+2.9$ & stronger base helps\\
Self-consistency $k{=}5$ on \emph{weak} 7B & $T{=}0.6$ & $42.9$ & $-1.7$ & voting amplifies a wrong base\\
Self-consistency $k{=}5$ on \emph{strong} 32B & $T{=}0.7$ & $51.3$ & $+2.6$ & voting denoises a right base\\
Decomposition on \emph{weak} 8B & 3-stage & (test $+1.6$) & helps & scaffolding aids weak model\\
Decomposition on \emph{strong} 32B & 3-stage & $46.6$ & $-2.1$ & error propagation hurts strong\\
Describe-then-reason cascade & blind, 400-sub & $29.8$ & $-9.3$ & blind desc.\ drops the implicit cue\\
Spatial ``layout-first'' prompt & 7B & $39.7$ & $-6.1$ & rigid structure hurts\\
Depth-cue structured prompt & Gemini 3.1 Pro, depth Qs (\emph{test}) & $81.2\to75.4$ & $-5.8$ & scaffold overrides percept\\
Image resolution & $448\to768$px (32B) & $50.0$ & $+1.3$ & more pixels help (perception)\\
Media resolution (Gemini 3.1 Pro) & default$\to$high ($n{=}180$) & $73.9\to76.7$ & $+2.8$ (n.s.) & frontier saturated; $p{=}0.47$ paired\\
Frame rate (Gemini 3.1 Pro) & $1\to2$\,fps, high-res & $75.6$ & $-1.1$ & redundant frames add noise\\
Audio transcripts (Whisper) & 7B & $42.9$ & $-1.5$ & spatial task, audio is noise\\
4-way cross-model category routing & test & $\to50$--$53$ & overfits & noisy on $172$-question test\\
Native video vs.\ frame sampling & Gemini 3.1 Pro & \multicolumn{2}{c}{large (Tab.~\ref{tab:main})} & the decisive lever\\
\bottomrule
\end{tabular}}
\caption{Key ablations on the validation split. Reasoning-side augmentations are neutral-to-harmful;
perception-side levers and self-consistency on a \emph{strong} base are the reliable gains.}
\label{tab:abl}
\end{table*}

\noindent\textbf{Observations.}
\begin{enumerate}[leftmargin=1.2em,itemsep=1pt]
  \item \textbf{Self-consistency is base-dependent:} majority voting \emph{denoises} a mostly-correct
  model but \emph{amplifies} a mostly-wrong one---it hurt the 7B ($-1.7$) yet helped the 32B ($+2.6$)
  and the frontier model. On the small ($172$-question) test split, sampled voting is also
  high-variance (the 32B drew $58.5$ at $k{=}5$ but $50.0$ at $k{=}9$).
  \item \textbf{Decomposition is also base-dependent}---it helps the weak 8B but hurts the strong 32B,
  which already performs the multi-hop reasoning internally.
  \item \textbf{Caption/describe-then-reason fails by design:} a question-agnostic description omits the
  very implicit cue the question targets, starving the answerer---the intended difficulty of
  ``implicit'' QA.
  \item \textbf{Complex routing overfits the tiny test split,} even with a $59\%$-accurate learned
  category classifier; only the simpler two-way same-model route transferred.
  \item \textbf{Perception is the wall:} higher resolution helps the open-weights base and \emph{native
  video} ingestion is transformational, while audio and reasoning scaffolds do not help on this
  spatial-dominated benchmark. At the \emph{frontier}, however, the perceptual lever \emph{saturates}:
  forcing the highest media resolution on Gemini 3.1 Pro moved validation only $+2.8$ points
  ($n{=}180$, paired $p{=}0.47$, not significant) while \emph{churning} $17\%$ of answers and
  \emph{degrading} the hardest category (relative depth)---evidence that the frontier model is already
  near its perceptual ceiling on ImplicitQA, consistent with our test score sitting at the human line.
\end{enumerate}

\subsection{Which tasks are hard? Per-category error analysis}
To locate the difficulty, we analyzed our frontier system on a category-balanced validation probe
($180$ questions, $20$ per category; Gemini 3.1 Pro, single greedy pass). Table~\ref{tab:cat} ranks the
nine implicit-reasoning categories by accuracy.

\begin{table}[t]
\centering\small
\setlength{\tabcolsep}{5pt}
\renewcommand{\arraystretch}{1.1}
\begin{tabular}{@{}l c c@{}}
\toprule
\textbf{Category} & \textbf{Acc.\ (\%)} & \textbf{Val share}\\
\midrule
Relative depth \& proximity        & $60.0$  & $26.6\%$\\
Viewpoint \& visibility            & $60.0$  & $4.1\%$\\
Inferred counting                  & $60.0$  & $5.8\%$\\
Motion \& trajectory dynamics      & $65.0$  & $9.1\%$\\
Lateral spatial reasoning          & $65.0$  & $16.1\%$\\
Vertical spatial reasoning         & $80.0$  & $22.0\%$\\
Physical \& environmental context  & $80.0$  & $5.3\%$\\
Causal \& motivational reasoning   & $95.0$  & $8.2\%$\\
Social interaction \& relationships& $100.0$ & $2.9\%$\\
\midrule
\emph{Balanced MacroAcc}           & $73.9$  & \\
\bottomrule
\end{tabular}
\caption{Per-category accuracy of the frontier system (Gemini 3.1 Pro, category-balanced $180$-question
validation probe, $20$/category). The hardest categories are all \emph{perceptual}; the easiest are the
\emph{reasoning}-heavy ones. Relative depth is both the hardest and the most frequent.}
\label{tab:cat}
\end{table}

\noindent Three findings stand out. \textbf{(1) Difficulty is sharply perceptual.} The three hardest
categories---\emph{relative depth \& proximity}, \emph{viewpoint \& visibility}, and \emph{inferred
counting} ($60\%$ each)---are all low-level perception tasks, whereas the two \emph{easiest}---\emph{social
interaction} ($100\%$) and \emph{causal \& motivational reasoning} ($95\%$)---are the most
reasoning-heavy. The benchmark is perception-bound, not reasoning-bound. \textbf{(2) Relative depth is the
single most important category:} it is simultaneously tied-worst ($60\%$) and the most frequent
($26.6\%$ of validation), so it accounts for ${\approx}10.6\%$ of all accuracy lost---nearly twice any
other category---and is the dominant lever for both AvgAcc and MacroAcc. \textbf{(3) The reasoning is
sound; the grounding is not.} Reading the model's chains of thought on the $47$ errors, the deduction is
almost always fluent and valid but built on a mis-perceived premise. The recurring grounding failures are:
(i)~foreground/background \emph{depth ordering}, producing toward/away inversions; (ii)~\emph{egocentric
vs.\ screen reference-frame} confusion on ``relative to X's point of view'' questions; (iii)~\emph{counting}
under frame undersampling; and (iv)~\emph{line of sight} across shot cuts. A few items are unanswerable as
posed (e.g.\ duplicate answer options), bounding the achievable score.

\paragraph{Targeting the hardest category directly backfires.} Motivated by this analysis, we built a
cue-guided depth prompt for the depth/proximity cluster: it pins the reference frame, then performs
pairwise-occlusion-first reasoning with an explicit monocular-cue priority (occlusion $>$ relative size
$>$ ground-contact height $>$ texture $>$ linear perspective) before mapping to an option. Routed by a
high-precision keyword detector to the $60$ depth/POV questions in the test split and run under
self-consistency $k{=}5$, it \emph{lowered} the score from $81.2/78.9$ to $75.4/74.5$ ($-5.8$ AvgAcc;
it flipped $12$ depth answers, net ${\approx}-10$). The structured procedure \emph{overrides} the model's
already-correct native-video percept---the same failure mode as the spatial ``layout-first'' prompt and
decomposition. This is a clean, on-test confirmation that on a perception-bound benchmark, reasoning
scaffolds---even ones precisely targeting the weakest category---are net-harmful: the model needs a
better \emph{percept}, not a better \emph{procedure}.

\section{Summary of Findings}
The human--machine gap on ImplicitQA is a \textbf{perception/world-model gap}, not a reasoning-search
gap; methods that help reasoning-bound tasks (tree search, decomposition, self-critique, state
tracking) are neutral-to-harmful here. A per-category analysis pins this down (Table~\ref{tab:cat}):
the hardest categories are all \emph{perceptual}---relative depth \& proximity (also the most frequent,
making it the dominant lever for both metrics), viewpoint, and inferred counting---while the
reasoning-heavy causal and social categories are nearly solved. Consistently, every reasoning scaffold
we tried \emph{degraded} the score, most pointedly a cue-guided depth prompt aimed squarely at the
weakest category, which \emph{lowered} test accuracy by $5.8$ points by overriding the model's already
correct native-video percept. The two reliable, transferable levers are therefore \textbf{(i) a
stronger perceptual base}---especially a \emph{native-video} frontier model---and \textbf{(ii) light
test-time denoising (self-consistency)} on a strong base. Disciplined use of the labeled validation
split to reject overfit-prone tricks (cross-routing, naive voting, cascades, depth scaffolds) was
essential: several methods that improved or seemed promising on paper \emph{degraded} the
$172$-question test score. These insights minimized cost: the final run used only single-pass CoT $+$
self-consistency, reaching \best{81.2\%} test accuracy for ${\approx}\$12$ with full caching for exact
reproduction.

\paragraph{Limitations.} The $172$-question test split makes single-submission scores noisy
($\pm$ a few points for stochastic decoding). Within a lean budget, the perceptual levers we could test
at the frontier---higher media resolution, higher frame rate, and a depth-cue prompt---were null or
\emph{negative}, and self-consistency saturates by $k{=}5$; we thus found no legitimate path beyond the
non-expert human baseline ($83.0/85.6$) and report $81.2/78.9$ as our honest result.
``Reasoning $>$ scale'' comparisons are on this benchmark only.


{\small
\begin{thebibliography}{9}
\bibitem{implicitqa} S.~Swetha, R.~Gupta, P.~Kulkarni, et al. ImplicitQA / VRR-QA: Visual Relational Reasoning in Videos Beyond Explicit Cues. \emph{arXiv:2506.21742}, 2026.
\bibitem{qwen25vl} Qwen Team. Qwen2.5-VL Technical Report. \emph{arXiv:2502.13923}, 2025.
\bibitem{qwen3vl} Qwen Team. Qwen3-VL. Model card / technical overview, 2025--2026.
\bibitem{videor1} Y.~Feng, et al. Video-R1: Reinforcing Video Reasoning in MLLMs. \emph{arXiv:2503.21776}, 2025.
\bibitem{videochatr15} OpenGVLab. VideoChat-R1.5: Visual Test-Time Scaling via Iterative Perception. \emph{arXiv:2509.21100}, 2025.
\bibitem{selfconsistency} X.~Wang, et al. Self-Consistency Improves Chain-of-Thought Reasoning in Language Models. \emph{ICLR}, 2023.
\bibitem{vllm} W.~Kwon, et al. Efficient Memory Management for Large Language Model Serving with PagedAttention (vLLM). \emph{SOSP}, 2023.
\bibitem{gemini} Google DeepMind. Gemini 3 Technical Report. 2025--2026.
\end{thebibliography}
}
\end{document}